%% file: main.tex
\pdfoutput=1

\documentclass[11pt]{article}

\usepackage[]{emnlp2022}

\usepackage{algorithm}
\usepackage{algpseudocode}
\usepackage{times}
\usepackage{latexsym}
\usepackage{cleveref}
\usepackage{graphicx}
\usepackage{multirow}
\usepackage{colortbl}
\usepackage{subcaption}
\usepackage{comment}
\usepackage{xspace}
\usepackage{booktabs}
\usepackage{inconsolata}
\usepackage{enumitem}
\definecolor{ForestGreen}{RGB}{34,139,34}

\usepackage[T1]{fontenc}

\usepackage[utf8]{inputenc}
\input{macros}
\algnewcommand\algorithmicforeach{\textbf{for each}}
\algdef{S}[FOR]{ForEach}[1]{\algorithmicforeach\ #1\ \algorithmicdo}
\usepackage{microtype}
\usepackage{multirow}

\title{\textit{Help me write a poem}: Instruction Tuning as a Vehicle for Collaborative Poetry Writing}

\author{Tuhin Chakrabarty$^1$\thanks{~~Both Authors Contributed Equally}~~~~~Vishakh Padmakumar$^{2*}$~~~~~He He $^{2,3}$ \\
 $^1$Department of Computer Science, Columbia University\\  
 $^2$Center for Data Science, New York University\\
  $^3$Department of Computer Science, New York University\\
 {\tt\small tuhin.chakr@cs.columbia.edu, vishakh@nyu.edu, hhe@nyu.edu} \\
}

\begin{document}
\maketitle
\begin{abstract}
Recent work in training large language models (LLMs) to follow natural language instructions has opened up exciting opportunities for natural language interface design. Building on the prior success of LLMs in the realm of computer-assisted creativity, we aim to study if LLMs can improve the quality of user-generated content through collaboration. We present \textit{CoPoet}, a collaborative poetry writing system. In contrast to auto-completing a user's text, CoPoet is controlled by user instructions that specify the attributes of the desired text, such as \textit{Write a sentence about `love'} or \textit{Write a sentence ending in `fly'}. The core component of our system is a language model fine-tuned on a diverse collection of instructions for poetry writing. Our model is not only competitive with publicly available LLMs trained on instructions (InstructGPT), but is also capable of satisfying unseen compositional instructions. A study with 15 qualified crowdworkers shows that users successfully write poems with CoPoet on diverse topics ranging from \textit{Monarchy} to \textit{Climate change}. Further, the collaboratively written poems are preferred by third-party evaluators over those written without the system.\footnote{Our code, preprocessed data, models, and the interaction logs from our user study are available at \url{https://github.com/vishakhpk/creative-instructions}}
\end{abstract}

\input{intro}

\input{data}

\input{experiments}
\input{results}

\input{copoet}
\input{related}
\input{conclusion}

                                                     \section*{Acknowledgements}
We would like to thank the anonymous reviewers for their helpful comments. We additionally also want to acknowledge all human authors who posted their work open-sourced on the websites we collected the data from. Tuhin is funded by Columbia Center of Artifical Intelligence \& Technology (CAIT) and the Amazon Science Ph.D. Fellowship. This work is also supported by the Samsung Advanced Institute of Technology (Next Generation Deep Learning: From Pattern Recognition to AI), the National Science Foundation under Grant No. 1922658, and a gift from AWS AI. 

\bibliographystyle{acl_natbib}
\bibliography{custom, all}

\appendix
\input{appendix}

\end{document}

%% file: macros.tex
\usepackage{array}
\newcommand{\PreserveBackslash}[1]{\let\temp=\\#1\let\\=\temp}
\newcolumntype{C}[1]{>{\PreserveBackslash\centering}p{#1}}
\newcolumntype{R}[1]{>{\PreserveBackslash\raggedleft}p{#1}}
\newcolumntype{L}[1]{>{\PreserveBackslash\raggedright}p{#1}}

\newcommand{\eg}{e.g.,\xspace}

\newcommand{\copoet}{\textit{CoPoet}\xspace}


%% file: intro.tex
\section{Introduction} \label{sec:intro}

Advancements in large language models (LLMs) have made remarkable progress towards generating coherent text in a wide variety of domains. This has spurred increasing interest in computer-assisted creativity \cite{see2019massively,elkins2020can,ramesh2022hierarchical,branwen2020gpt} such as building co-creative assistants for writing stories, poems, and argumentative essays \cite{lee2022coauthor,swanson2021story,uthus2019first,donahue2020enabling,padmakumar2021machine,du-etal-2022-read}. The adoption of these technologies hinges on their ability to provide appropriate suggestions while being easy to interact with. However, there has been limited research on the effectiveness of such collaboration, \eg whether the assistant understands user intents and whether collaboration improves the final outcome.

\definecolor{accept}{HTML}{CCE5FF}
\definecolor{human}{HTML}{D5E8D4}
\definecolor{reject}{HTML}{FFCCCC}

\begin{figure}
    \centering
    \includegraphics[width=0.475\textwidth]{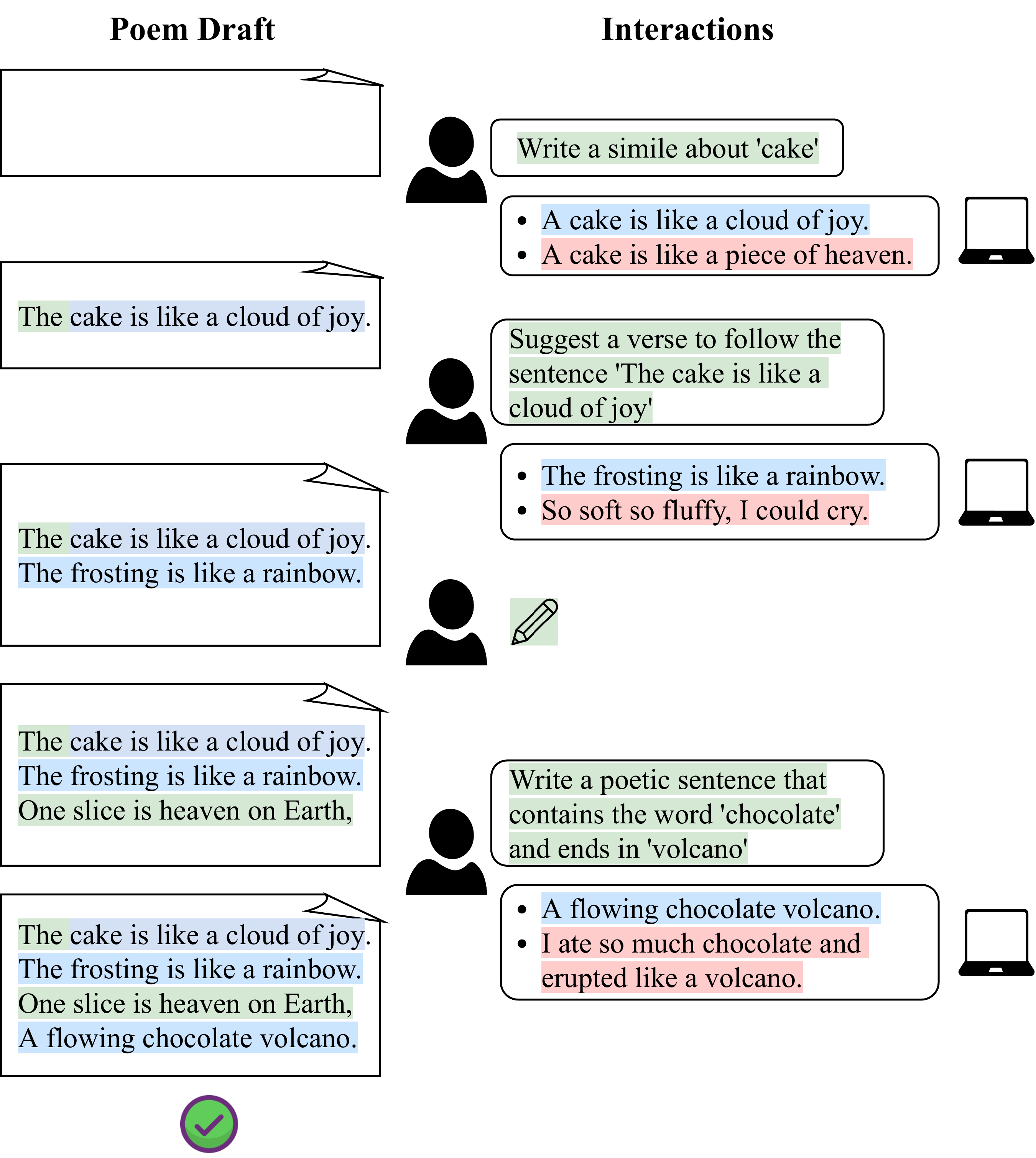}
    \caption{
    A collaborative poem entitled '\textit{Decadence}', written with CoPoet assistance. \colorbox{human}{Green text} was written directly by the human, who interacts with CoPoet using instructions. CoPoet offers multiple suggestions which the user can 
    \colorbox{accept}{accept} 
    or \colorbox{reject}{reject}.
    The user wrote a four line poem before indicating completion of the task.
    }
    
    \label{fig:copoet_overview}
\end{figure}

In this paper, \textit{we aim to understand the collaboration capabilities of LLMs through a case study of collaborative poetry writing.} 
Writing a poem is often a challenging task because it is both open-ended and highly constrained. Unlike stories or other argumentative texts, in order to write a poem we need creative content that satisfies various long- and short-range form constraints such as rhyme, meter, and sound, 
which poses a significant challenge for end-to-end poem generation systems \cite{ghazvininejad-etal-2016-generating,tian-peng-2022-zero,van2020automatic,ormazabal2022poelm}. While LLMs sometimes struggle with long-range coherence, they are good at providing variations of text that satisfy local 
constraints.
This makes them great partners to humans in poem writing, 
where humans focus on the long-range writing plan and the machine implements the ideas locally.

Effective collaboration in co-creative writing is challenging as it requires the model to understand user intention.
For example, as shown in Figure \ref{fig:copoet_overview},
a user may have a rough plan around two related concepts such as {chocolate} and volcano,
and want the model to suggest a verse that contains \textit{chocolate} and ends with \textit{volcano};
or they may be looking for a verse that rhymes with a specific word (\textit{rainbow}) to satisfy the constraints.
An auto-completion interface is not able to 
anticipate such 
user needs and provide targeted suggestions.
To enable richer interaction,
we rely on instructional prompts \cite{wang2022benchmarking, sanh2021multitask,mishra-etal-2022-cross,mishra2022help} that act as a natural language interface between the user and the assistant.

Specifically, we present \textit{CoPoet}, a collaborative poem writing system with a natural language interface.
During a writing session, the user can iteratively request suggestions through natural language instructions such as \textit{Write a simile about `cake'},
and edit their draft based on the suggestions
(\Cref{fig:copoet_overview}).
To build \copoet, we finetune a pretrained sequence-to-sequence model on a parallel corpus of instruction-output pairs.
We obtain the outputs from publicly available datasets of creative text
and synthesize the corresponding instructions by rules,
including both lexical and rhyming constraints as well as requests on rhetorical devices.

To understand how well the model follows instructions,
we test it on instructions with varying levels of difficulty,
from those seen during training to unseen compositional instructions that contain 
multiple constraints.
Both automatic and human evaluation show that our finetuned  
model satisfies the constraints 86\% of the time, 10\% better than a much larger 175B version of InstructGPT \cite{brown2020gpt3}. 
On unseen compositional instructions, our best model satisfies them 77.6\% of the time, outperforming InstructGPT by a margin of 28\%.

To understand its collaboration capabilities,
we run a user study on Amazon Mechanical Turk (AMT) where \copoet assists expert crowd workers (recruited through a qualification test) in writing poems (\Cref{co-poet-description}). We observe that the recruited users are able to write coherent and creative poems on diverse topics ranging from \textit{Glass Ceiling} to \textit{Climate Change}.
About 70\% of model suggested text is retained in the final poem and users give \copoet a rating of 4.3 out of 5 on both the suggestion quality and the overall helpfulness. Further, a separate group of annotators on AMT prefers the collaboratively written poems more often than those written without \copoet assistance.
In particular, we find model assistance improves rhyming and vocabulary diversity of the poems.

%% file: data.tex
\section{Data}
\label{sec:train_data}


To train a model to follow instructions
, we need \texttt{<instruction, poem\_line>} 
pairs where the text satisfies 
the instruction. The key challenge to building such a model is the lack of parallel data, so we collect our own dataset of creative writing instructions 
from publicly available poem corpora or relevant subreddits from Reddit 
(Table \ref{instrsource}).

\begin{table}[]
\centering
\small
\renewcommand{\arraystretch}{1.25}
\begin{tabular}{|p{1.05cm}|p{5.5cm}|}
\hline
\multirow{2}{*}{Subject}                                                  & Write a poetic sentence about {\color{ForestGreen}`sun'}                                                                               \\ \cline{2-2} 
                                                                          & \textit{O crimson {\color{blue}sun}, your warming draft's pulsation.}                                                                    \\ \hline\hline
\multirow{2}{*}{End}                                                      & Write a poetic sentence ending in {\color{ForestGreen}`glory'}                                                                         \\ \cline{2-2} 
                                                                          & \textit{Am I exalted here unto that {\color{blue}glory}.}                                                                               \\ \hline\hline
\multirow{2}{*}{Rhyme}                                                    & \begin{tabular}[c]{@{}l@{}}Write a poetic sentence that ends in a word\\ which rhymes with {\color{ForestGreen}`replace'}\end{tabular} \\ \cline{2-2} 
                                                                          & \textit{Diminishing for me, with delicate {\color{blue}grace.}}                                                                       \\ \hline\hline
 \multirow{3}{*}{\begin{tabular}[c]{@{}l@{}}Next\\Sentence\end{tabular}} & \begin{tabular}[c]{@{}l@{}}Write a next sentence in a poem given the \\previous sentence `{\color{ForestGreen}`The only thing I know'} \\{\color{ForestGreen}`for sure'}\end{tabular}        \\ \cline{2-2} 
                                                                          & \textit{\color{blue}Is that love is hard and can be obscure.}                                                                     \\ \hline\hline                                                                         
\multirow{2}{*}{Metaphor}                                                 & Write a metaphor about {\color{ForestGreen}`brain'}                                                                                \\ \cline{2-2} 
                                                                          & \textit{My brain is a {\color{blue}tangled mess of circuits.}}                                                                         \\ \hline\hline
\multirow{2}{*}{Simile}                                                   & \begin{tabular}[c]{@{}l@{}}Write a simile about {\color{ForestGreen}`making someone feel}\\ {\color{ForestGreen}desired'}\end{tabular}                      \\ \cline{2-2} 
                                                                          & \begin{tabular}[c]{@{}l@{}}{\textit{I want to make you feel {\color{blue}like a flower near}}}\\ {\color{blue}\textit{a hummingbird}}\end{tabular}                \\ \hline\hline
\multirow{2}{*}{\begin{tabular}[c]{@{}l@{}}Onoma-\\ topoeia\end{tabular}} & \begin{tabular}[c]{@{}l@{}}Write a poetic sentence about {\color{blue}`bottles'} \\ showcasing onomatopoeia\end{tabular}        \\ \cline{2-2} 
                                                                          & \textit{The stamping of feet and the \color{blue}{ring of bottles.}}                                                                     \\ \hline\hline
\multirow{2}{*}{\begin{tabular}[c]{@{}l@{}}Subject\\+ End\end{tabular}} & \begin{tabular}[c]{@{}l@{}}Write a poetic sentence about {\color{ForestGreen}`tears'} and\\ ending in {\color{ForestGreen}`wives'}\end{tabular}        \\ \cline{2-2} 
                                                                          & \textit{Awash in the {\color{blue}tears} of soldier's {\color{blue}wives.}}                                                                   \\ \hline                                                                          
\end{tabular}
\caption{\label{trainins}Natural language instructions for poem writing paired with example outputs.
Each instruction consists of a template and an \textcolor{ForestGreen}{argument}. 
}
\vspace{-3ex}
\end{table}

Based on some initial feedback from professional poets, we decided to include 3 major types of instructions: 1) \textit{Continuation} based instructions that suggest content when writers are blocked/clueless on how to proceed; 2) Instructions on \textit{Lexical Constraints} to enable greater control of poetic form such as rhyme, sound, and meter. These are instructions that force language models to obey specific choices such as generating a line  
that contains a specific \textit{topic}, \textit{start word}, \textit{end word} or a sentence with a particular \textit{rhyme}; 3) Instructions on \textit{Rhetorical devices} that are mostly used for introducing embellishments and imagery in a poem such as \textit{metaphor, similes, and onomatopoeia}. 

Table \ref{trainins} shows the primary instructions used to train our models. These instructions are crafted by the authors of the paper, who convert every poem line to an \texttt{<instruction, poem\_line>} 
pair using rules.




Each instruction consists of a \emph{template} (unique to the instruction type) and one or more \emph{arguments}, as can be seen in Table \ref{trainins}.
Given a poem line in the corpus, we reverse-engineer the instruction by picking a template and extracting the arguments from the poem line. 
For continuation instructions, we use the previous context as the argument. For instructions on lexical constraints, we extract noun phrases and start/end words as arguments using NLTK for tokenization. To construct instructions on rhymes, we use the CMU dictionary to find rhyming words.\footnote{\url{https://pypi.org/project/pronouncing/}}
We describe more details in Appendix \ref{sec:appendix} on how we create instructions for each particular type.

To allow models to adapt to linguistic variations of the instruction templates, we also include paraphrases of the instruction templates, e.g., instead of ``Write" we also use``Generate'', or instead of ``Write a sentence about'' we use ``Write a sentence that contains the word'' or ``Write a sentence that includes the word''. In total, our dataset consists of 873,574 \texttt{<instruction, poem\_line>} pairs which we randomly split into 808,180 train and 65,394 held-out validation examples.\footnote{Our dataset is publicly available at \url{https://github.com/vishakhpk/creative-instructions}.} We evaluate performance on three test sets of hand-crafted instructions of varying difficulty (\Cref{sec:test_sets}).

%% file: experiments.tex
\section{How Well Do LLMs Follow  Instructions?}
\label{sec:instruction_tuning}
In this section, 
we first describe our models and baselines, followed by 
the evaluation results using both automatic metrics (\Cref{sec:automatic_evaluation}) and human evaluation (\Cref{sec:human_evaluation}).

\subsection{Experiment Setup}
\paragraph{Model Details}
We finetune 
the pretrained T5 \cite{raffel2019exploring} and T0 \cite{sanh2021multitask} models from HuggingFace \cite{wolf2019transformers} on the collected data (\Cref{sec:train_data}) to produce the output given the instruction using cross-entropy loss.  We report results on finetuned T5-3B, T5-11B and T0-3B models, which are henceforth referred to as T5-3B-poem, T5-11B-poem, and T0-3B-poem. 
We select the hyperparameters by the validation loss:
for T5-11B-poem, we use the Adam optimizer with a learning rate of $1e^{-4}$;
for T5-3B-poem and T0-3B-poem, we use the Adafactor optimizer with a learning rate of $1e^{-3}$. 
Each model is trained for 3 epochs with early stopping based on validation loss.
We finetune all models on an A100 GPU and use Deepspeed \cite{rasley2020deepspeed} integration for the 11B model. During finetuning, we restrict the maximum sequence length of both the source and the target to 64 tokens (via truncation).\footnote{The length limit is chosen to avoid memory explosion. It has minimal impact on model performance since most verses are shorter.} 
At inference time, we generate output sequences using top-k sampling with $k=5$ and a temperature of $0.7$ per recommendations from earlier work in open-ended creative text generation \cite{fan-etal-2018-hierarchical, holtzman2020curious, padmakumar2021machine}.

\paragraph{Baselines}

We compare our finetuned models 
with two other models:
(i) the T0pp model \cite{sanh2021multitask}, trained on instruction-based prompts from 49 datasets;\footnote{These include question-answering, summarization, structure-to-text generation, sentiment and topic classification tasks but no explicit creative writing tasks.}
and (ii) the 175B davinci variant of InstructGPT \cite{ouyang2022training} that is trained on human-written instructions on diverse tasks in a human-in-the-loop fashion.
Given an instruction, we generate text directly (i.e.\ zero-shot) from T0pp using top-k sampling \cite{fan-etal-2018-hierarchical}.

For InstructGPT, we evaluate on 
both zero-shot and few-shot settings. 
For zero-shot, the prompt consists of only the instruction. 
For few-shot, the prompt consists of 26 \texttt{<instruction, poem\_line>} pairs 
from our training data (selected to 
cover all the instruction templates), 
followed by the test instruction.\footnote{The exact prompt can be found in our code repository.}
We use the OpenAI API with a temperature of $0.7$, no frequency penalty, and a maximum sequence length of 64 to match our setting.

\subsection{Test Sets}
\label{sec:test_sets}

While our training instructions cover many templates and topics,
user instructions may deviate from the training distribution during interaction.
To evaluate the generalization capabilities of the  
models,
we identify three settings with increasing difficulty based on whether the instruction templates or 
arguments are seen during training.

\paragraph{Known Instruction Templates with Known Arguments (KIKA)}
The simplest setting requires the model to generalize to novel combinations of the templates and arguments.
Specifically, we create instructions where
both the templates and the arguments are seen in the training set, although each specific combination is unseen (i.e.\ the training and test sets have no overlapping instructions). 

\paragraph{Known Instruction Templates with Unknown Arguments (KIUA)}
To handle novel concepts from users, the model must generalize to unseen arguments,
which may include new entities or phrases. 
For example, it might be easier for a model to write a poetic sentence about a known argument such as \textit{beauty}, but difficult to write about an unknown argument \textit{beauty without virtue}.
For this set, we include instructions where the instruction templates are seen during training but the 
corresponding arguments are unseen. 

\paragraph{Unknown Compositional Instruction Templates} 
One of the main benefits of natural language instructions is that they can be easily composed in new ways to cover various user intentions. This is particularly useful in creative writing because it enables users to request text from the model with multiple constraints. Therefore, we also test whether the model understands compositional instructions using two templates, as seen in \Cref{testcomp}. 
Our model is exposed to a \emph{single} compositional template during training: \textit{Subject+End}. For this test set, we create a variety of unseen compositions. 

In total, we create 242 test examples (82 KIKA, 82 KIUA, 78 compositional) by selecting instructions according to the above criteria, followed by manual verification.

\begin{table}[]
\centering
\small
\renewcommand{\arraystretch}{1.45}
\begin{tabular}{|p{1.05cm}|p{5.5cm}|}
\hline
\multirow{2}{*}{\begin{tabular}[c]{@{}l@{}}Start\\ +End\end{tabular}}           & \begin{tabular}[c]{@{}l@{}}Write a poetic sentence that starts with the \\ word {\color{ForestGreen}`Maybe'} and ending in {\color{ForestGreen}`void'}\end{tabular}                                               \\ \cline{2-2} 
                                                                                & \textit{\color{blue}Maybe one day, you will find me in the void}                                                                                                                            \\ \hline\hline
\multirow{2}{*}{\begin{tabular}[c]{@{}l@{}}Subject\\ +Rhyme\end{tabular}}       & \begin{tabular}[c]{@{}l@{}}Write a poetic sentence that contains the \\ word {\color{ForestGreen}`breaks'} and ending in a word which\\ rhymes with {\color{ForestGreen}`bound'}\end{tabular}                     \\ \cline{2-2} 
                                                                                & \textit{\color{blue}She cracks and breaks and hits the ground.}                                                                                                                        \\ \hline\hline
\multirow{2}{*}{\begin{tabular}[c]{@{}l@{}}Next\\ Sentence\\ +End\end{tabular}} & \begin{tabular}[c]{@{}l@{}}Write a next sentence in a poetry given \\ the previous sentence {\color{ForestGreen}`Every once a while}\\ {\color{ForestGreen}I lower the blinds'} and ending in {\color{blue}`play'}\end{tabular} \\ \cline{2-2} 
                                                                                & \textit{\color{blue}Waiting for someone to call me out to play}                                                                                                                             \\ \hline\hline
\multirow{2}{*}{\begin{tabular}[c]{@{}l@{}}Metaphor\\ +End\end{tabular}}        & \begin{tabular}[c]{@{}l@{}}Write a metaphor that includes the word\\ {\color{ForestGreen}`film'} and ending in {\color{ForestGreen}`thought'}\end{tabular}                                                        \\ \cline{2-2} 
                                                                                & \textit{\color{blue}A film is a petrified fountain of thought.}                                                                                                                              \\ \hline
\end{tabular}
\caption{\label{testcomp}Examples of compositional natural language instructions for creative tasks paired with their respective outputs from our test sets.}
\end{table}

%% file: results.tex
\label{sec:results}



\subsection{Automatic Evaluation}
\label{sec:automatic_evaluation}
We evaluate how well the models satisfy constraints specified in the instructions on each of the test sets (\Cref{sec:test_sets}). 
We 
report the success rate of satisfying the instructions where the success condition for each instruction type is listed in 
\Cref{tab:success_conditions}.\footnote{Prior work on instruction tuning reports metrics such as BLEU score for generation tasks \cite{sanh2021multitask, wei2021finetuned} and these are unsuitable for our poetry writing instructions, thus we define custom success conditions.} 

\begin{table}[ht!]
\centering
\small
\def\arraystretch{1.35}
\begin{tabular}{| L{1.35cm} | L{5.25cm} |} 
\hline
\textbf{Instruction Type} & \textbf{Success Condition} \\ 
\hline
Rhyme& Last word of the model generation rhymes with the desired subject using the CMU Pronouncing Dictionary  \\
\hline
Haiku& Model generation contains 15--19 syllables and contains the desired subject  \\
\hline
Simile / Metaphor  & Model generation contains the desired subject as well as a comparator   \\
\hline
Start / End& First/last word of the model generation matches the desired subject \\
\hline
Subject  & Model generation contains the desired subject in the instruction\\
\hline
\end{tabular}
\caption{Success conditions for different instruction templates.}
\label{tab:success_conditions}
\vspace{-0.5 cm}
\end{table}

\begin{figure}
 \includegraphics[width=0.45\textwidth]{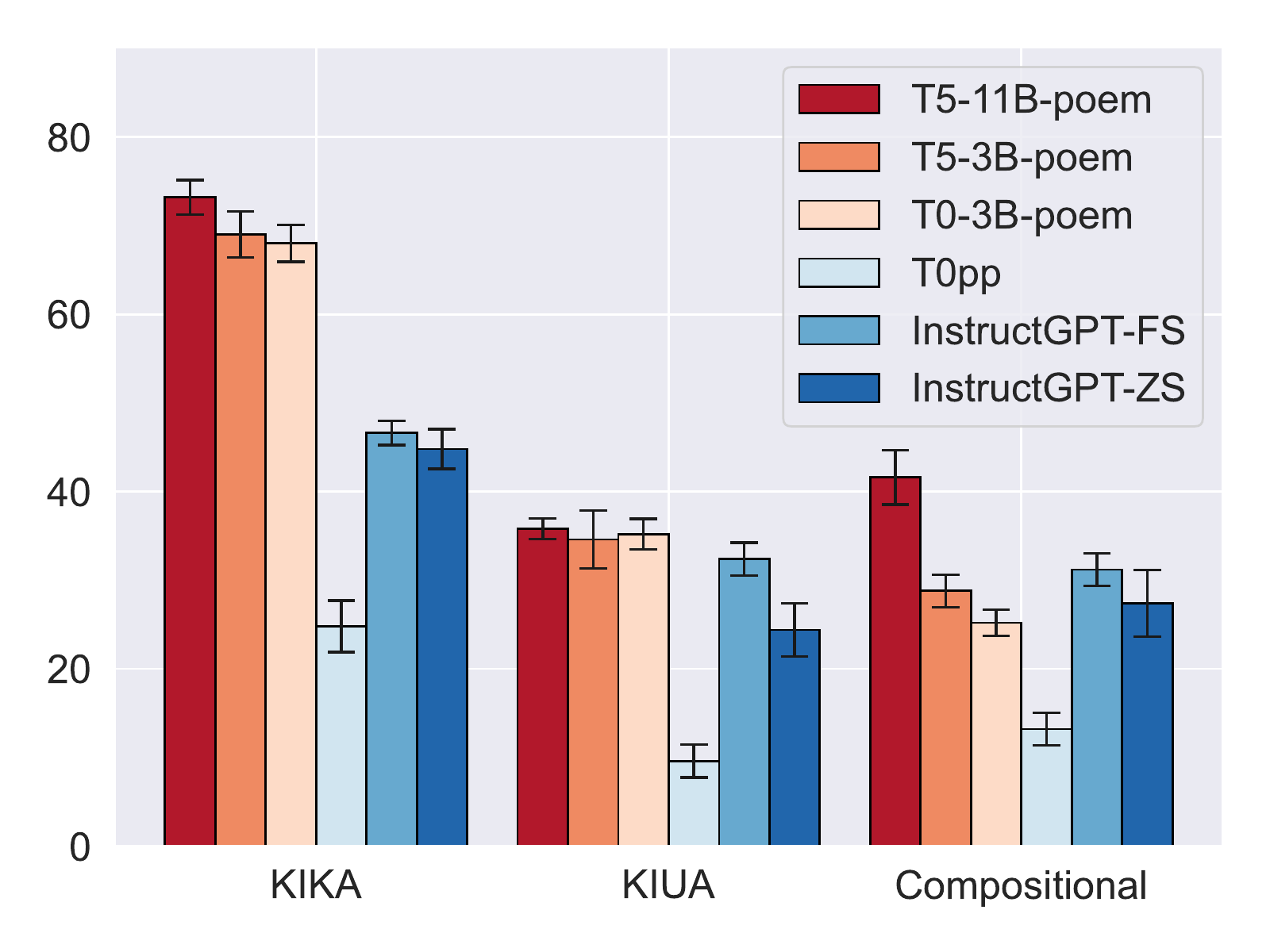}
 \label{fig:kike_automatic}
\caption{Automatic evaluation of models on KIKA, KIUA and Compositional test sets. The $y$ axis is the percentage of instructions that each model successfully satisfies as determined by the criteria in \Cref{tab:success_conditions}. We report results on T5-11B-poem, T5-3B-poem and T0-3B-poem along with the baselines---zero-shot T0pp \cite{sanh2021multitask} and zero-shot (ZS)/few-shot (FS) InstructGPT (da-vinci) \cite{ouyang2022training}. Each bar shows the average success rate of 5 model inferences along with the standard deviation. On average, T5-11B-poem achieves the highest success rate and InstructGPT is a strong few-shot baseline that obtains comparable results on KIUA.}
\label{fig:automatic_eval}
\end{figure}

\paragraph{Finetuned Models Have Strong In-Domain Performance but Drop on Out-of-Domain Data} 
\Cref{fig:automatic_eval} shows the average success rate and standard deviations of each model on the three test sets across 5 model inferences to account for variance in top-k sampling. On both KIKA and KIUA, T5-11B-poem has the highest average success rate. T5-3B-poem and T0-3B-poem outperform the few-shot and zero-shot baselines on both test sets. However, these finetuned models suffer a big drop in performance from KIKA to KIUA---T5-11B-poem suffers a relative drop of 51.09\% from a 73.2\% success rate on KIKA to a 35.8\% rate on KIUA. 
In contrast, the few-shot InstructGPT baseline only suffers a relative drop of 30.4\% from a success rate of 46.6\% on KIKA to 32.4\% on KIUA. 
This result is consistent with prior findings that  
task-specific finetuning may destroy pretrained representation which leads to degrading performance on other non-finetuning tasks \cite{aribandi2021ext5, padmakumar2022exploring}.
Without finetuning, in-domain examples are still helpful though: on all test sets, the InstructGPT few-shot baseline outperforms the corresponding zero-shot baseline along with a reduction in variance across runs. 

\paragraph{Larger Models Compose Instructions Better} 
On compositional instructions, we find that T5-11B-poem has the best average performance.
In addition, there is a clear performance gap between the 11B and 3B models,
showing the importance of model scale for composition, similar to recent observations of \textit{emergent} abilities in LLMs \cite{wei2022emergent}.
We also find that few-shot InstructGPT outperforms T5-3B-poem and T0-3B-poem despite having no compositional instructions in the prompt.
This indicates that smaller models, when finetuned on instructions, tend to overfit to templates seen during training, which hurts their generalization capability, as also reported in \citet{wei2021finetuned}.

\subsection{Human Evaluation}
\label{sec:human_evaluation}
Since our automatic metrics are not always accurate in measuring if an instruction is satisfied, 
we also perform human evaluation by having crowd workers manually check if model generations satisfy the instruction constraints.
Given the automatic evaluation results in \Cref{sec:automatic_evaluation}, we compare our best finetuned model, T5-11B-poem, against the top performing baseline, few-shot InstructGPT.
Specifically, we conduct pairwise comparison:
each annotator is shown an instruction and generations from both models.\footnote{We sample 5 generations from each model and select the best one using the criteria in \Cref{tab:success_conditions}. If multiple candidates are evaluated as success, we randomly sample one.} 
They are asked to rate the fluency, accuracy, and creativity of the generation by answering the following questions:
\begin{itemize}[itemsep=0pt, topsep=2pt, leftmargin=1em]
    \item Rate the fluency of each verse on a scale of 1--5.
    \item Does each verse adequately satisfy the instruction? (Yes/No) 
    \item Which of the two verses is \textit{more creative/interesting} while being \textit{coherent} and \textit{satisfying the instruction}? 
\end{itemize}
The first two questions evaluate the quality of each verse against the instruction individually. 
In addition to satisfying the constraints in the instruction with fluent text, we want the model to provide novel suggestions that are helpful for creative writing.
Thus we also ask the annotators to compare the two verses and provide a subjective judgement on which one is more creative. We collect three annotations for each question and use the majority vote as the final judgement. 

\begin{table}
\centering
\small
\begin{tabular}{clrr} 
\toprule
  &  & \textbf{T5-11B-poem} & \textbf{GPT3-FS}  \\ 
\midrule
\multirow{3}{*}{\textbf{KIKA (82) }} & Success\%& \textbf{86.2}  & 76.9 \\
  & Fluency  & 0.739  & \textbf{0.794}   \\
  & Creative  & \textbf{53.8}  & 46.2 \\ 
\midrule
\multirow{3}{*}{\textbf{KIUA (82) }} & Success\% & \textbf{92.5}  & 86.5 \\
  & Fluency  & 0.773 & \textbf{0.781}\\
  & Creative  & \textbf{56.7}  & 43.3 \\ 
\midrule
\multicolumn{1}{l}{}  & Success\% & \textbf{77.6}  & 55.2 \\
\multicolumn{1}{l}{\multirow{2}{*}{\textbf{Comp (78) }}} & Fluency  & 0.697  & \textbf{0.751}   \\
\multicolumn{1}{l}{}  & Creative  & 47.7   & \textbf{52.3}\\
\bottomrule
\end{tabular}
\caption{Human evaluation of model generations from T5-11B-poem and few-shot InstructGPT3 on different test sets across three metrics: (i) success rate: percentage of instructions satisfied; (ii) fluency: average fluency score on a scale of 5 normalized to $[0, 1]$; (iii) creativity: percentage of generations rated to be more creative / interesting in a pairwise comparison.
}
\label{tab:human_eval}
\end{table}

\paragraph{T5-11B-poem Satisfies Instructions Better than Few-Shot InstructGPT}
\Cref{tab:human_eval} shows the human evaluation results on all three test sets. We find that, on average, model generations from T5-11B-poem satisfy the given instructions better on all three test sets, while InstructGPT is rated to be more fluent consistently. 
We find that gap in satisfying instructions is largest on the compositional test set---T5-11B-poem accurately answers 77.6\% of compositional instructions while InstructGPT only manages 55.2\%. 
Annotators also reported that verses from T5-11B-poem were marginally more creative/interesting than InstructGPT on KIKA and KIUA test sets and less so on the Compositional test set,
indicating that the two models may have little difference in creativity.\footnote{The first two questions are less subjective than the third question. Users unanimously agreed 52.2\% of the time on whether model generations satisfied instructions and only 37.3\% on which output is more creative.} 


We observe that InstructGPT is a strong baseline, outperforming T0pp by a large margin on automatic metrics, and satisfying nearly 80\% of the instructions in the KIKA and KIUA test sets according to human evaluation. 
However, a common error case 
on compositional instructions is that while the model generations almost always contain the arguments mentioned in the instruction, they do not always satisfy the constraints correctly---when asked for a verse that contains the word `soul' and ends with `yellow', InstructGPT generated the line ``My soul is as yellow as the sun on a summer day'' that contains those arguments but not at the specified positions.  


\paragraph{Takeaways}
We observe that on average finetuned models tend to outperform the few-shot baselines on in-domain instructions (\Cref{sec:automatic_evaluation}). While smaller models (T5-3B-poem, T0-3B-poem) have worse performance on out-of-domain instructions, finetuned models at scale (T5-11B-poem) generalizes to compositional instructions effectively, even outperforming InstructGPT (\Cref{sec:human_evaluation}).
The flexibility of composing instructions makes the model more suitable as a collaborator for a human user; hence we use T5-11B-poem as the assistant for our subsequent collaborative experiments.

%% file: copoet.tex
\section{CoPoet: Collaborative Poem Writing}
\label{co-poet-description}

\begin{figure}[ht!]
    \centering
    \includegraphics[width=0.45\textwidth]{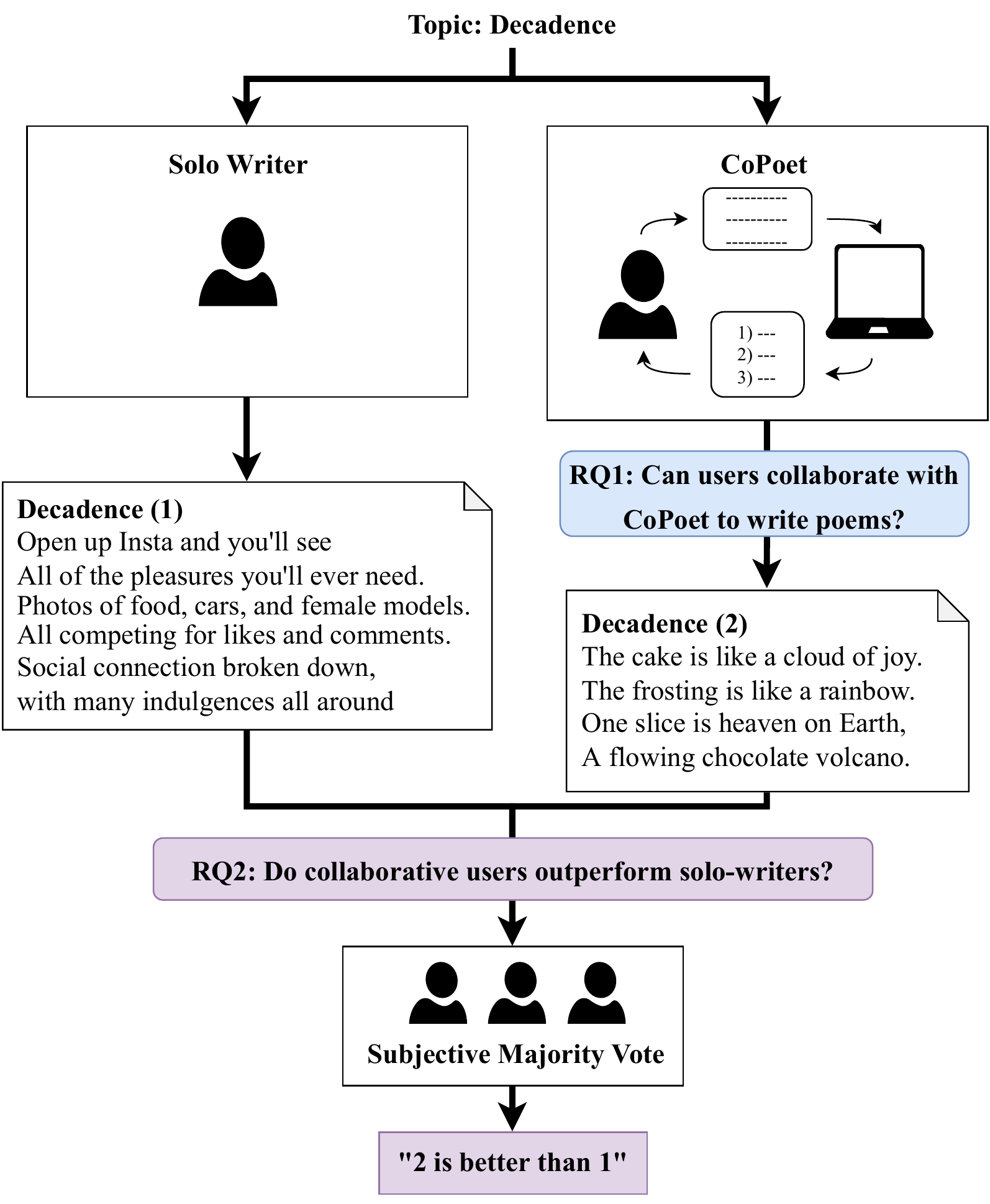}
    \caption{CoPoet user study. We study if users can effectively collaborate with CoPoet to write poems (RQ1) and whether writing with CoPoet produces better poems compared to solo-writers (RQ2). 
    }
    \label{fig:copoet_third_party}
\end{figure}

Our results in Section \ref{sec:human_evaluation}  demonstrate \copoet's ability to satisfy the constraints specified in the instructions. This presents us with an opportunity to test the model's capability in collaborative writing tasks. 
We design our user study (\Cref{fig:copoet_third_party}) to answer the following 
two main research questions:
\begin{itemize}
    \itemsep=0mm 
    \item \textbf{RQ1:} Can users write poems on any topic of their choice by collaborating with CoPoet?
    \item \textbf{RQ2:} Does CoPoet help users write \emph{better} poems compared to when they write alone? 
\end{itemize}

\paragraph{Interface Design}
Since we intend to study 
the task of collaborative poem writing, 
we develop a user interface for our experiments where users can work on their poem drafts and also query CoPoet for suggestions using written instructions. A screenshot of the interface is provided in \Cref{fig:architecture_copoet}. In response to each instruction, CoPoet provides 5 suggestions, each in the form of one poem line, to the users. The users can then choose if they wish to incorporate these into their draft.
We instruct them 
to edit the model output when required to ensure the overall coherence of the poem. 
As seen in \Cref{fig:architecture_copoet}, users are also provided with the list of instruction templates used to train the model (Section \ref{sec:train_data}). These are intended to communicate to users the instructions that the model is trained to respond to, so that they have an idea of what the model is capable of.\footnote{We explicitly mention that they can use novel instructions not present in the templates.}

\paragraph{Experiment Setup}
We first conduct a qualification test on AMT, where we recruit 50 workers 
to collaboratively write a poem of four lines 
using CoPoet. We require a user to interact with our system at least four times (i.e., to issue at least four instructions). 
However, we do not enforce that they 
use any of the model outputs in response to their 
instructions---they are free to ignore all model suggestions.
The authors of the paper then independently rank these poems in terms of fluency, richness in imagery, and creativity. Finally, 15 crowd-workers passed the qualification test. From now on, we refer to these qualified workers as \textit{experts}. 

We then 
collect 50 distinct poems collaboratively written by our experts using CoPoet, 
where they are instructed to write a poem on a topic of their choice. 
In order to compare collaborative writers to solo-writers, we then collect 50 poems on the same titles from expert writers writing without model assistance.\footnote{We ensure that the same author does not write on the same topic in the two setups.}
Third-party annotators were then shown the title and two poems interpreting it, and instructed to select the one they felt was a
‘descriptive interpretation of the title’. To ensure a fair judgement, both the poems were identical in length ($4$ lines), randomized in order, and without obvious clues in the vocabulary usage. To the best of our knowledge, there was no underlying bias that would make it easy for judges to identify which poems were collaborative and which were written entirely by humans. The full experiment design is shown in \Cref{fig:copoet_third_party}.

\paragraph{RQ1: Can experts write poems successfully on any topic of their choice by collaborating with CoPoet?\\}
From our user study, we observe that experts are able to collaborate with CoPoet and write poems on diverse topics of their choice, 
including 
\textit{Climate Change}, \textit{Hunger}, \textit{Glass Ceiling}, \textit{Decadence} etc. We include more examples in \Cref{sec:poem_examples}. The full list of titles 
visualized as a word cloud can be found in \Cref{fig:wcloud}. 

\begin{figure}[ht!]
    \centering
    \includegraphics[width=0.42\textwidth]{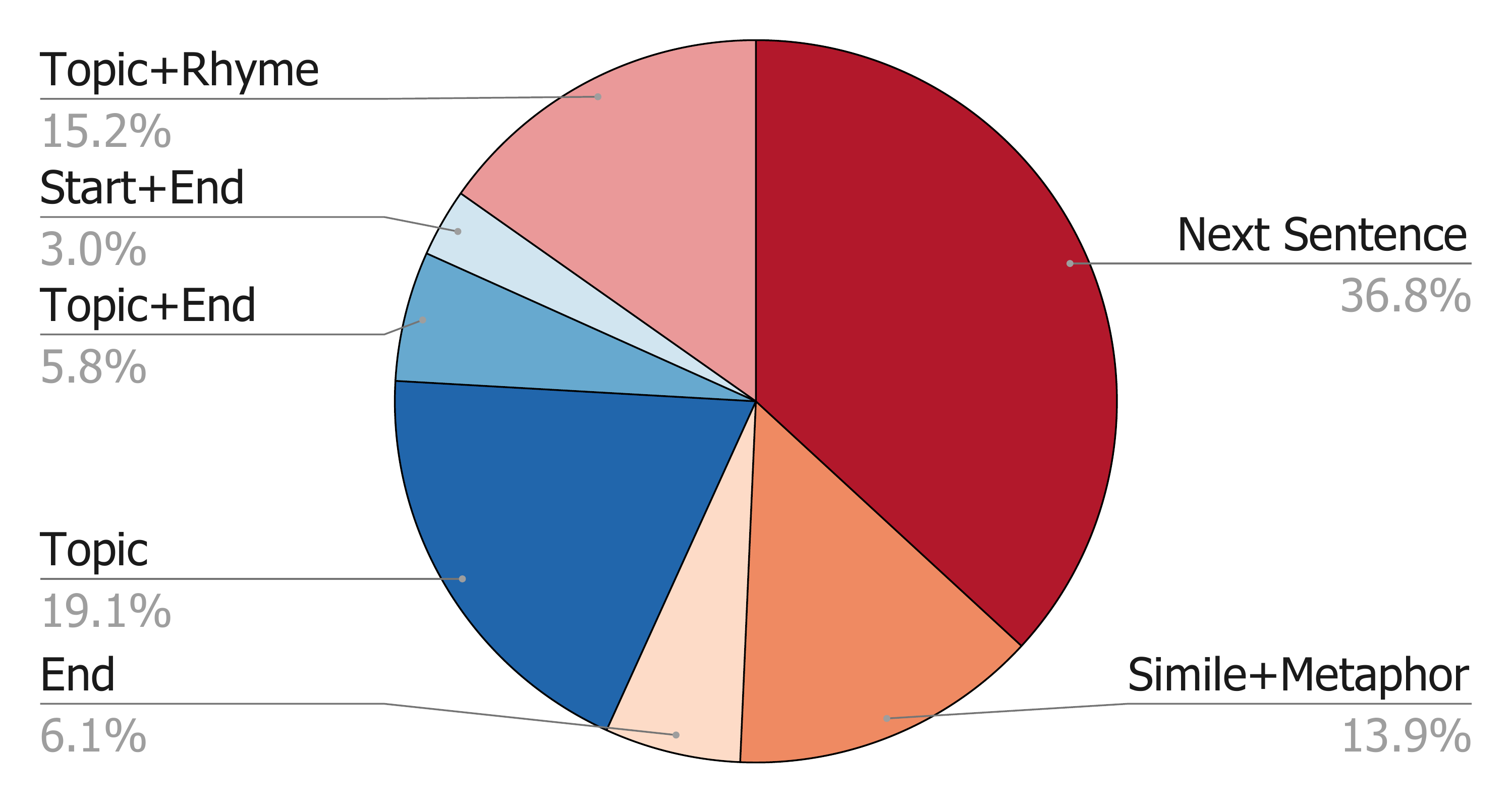}
    \caption{Proportions of the types of 
    instructions used by experts in the poetry writing task. 
    }
    \label{fig:inst_count}
\end{figure}

\paragraph{How do experts use instructions?}
On average, experts use 7 instructions per poem.
\Cref{fig:inst_count} shows that experts often prefer contextual instructions, i.e. getting ideas from the model about the \textit{Next Sentence} given what they have written thus far. The \textit{Topic} instruction is also significantly used, which helps them add control. It is encouraging to see humans using a total of 87 compositional instructions, which constitutes almost 24\% of the total set of instructions used. Finally, humans also use figurative embellishments such as \textit{Similes} or \textit{Metaphors} suggested by the model. 

\paragraph{Do experts find CoPoet helpful as a writing tool?}
We collect judgments from 15 experts to tease out and characterize the model's contribution. We are interested to know 
whether the model helped in the writing process by satisfying the instructions, and how well 
it served the writers' needs. 
We collect ratings on a Likert scale from 1 (not at all) to 5 (very) on two questions: (i) How accurately does 
the model follow instructions? 
(ii) How helpful is the model in the process of writing poetry? We obtain an average score of 4.3 out of 5 on both questions, suggesting that CoPoet is a useful tool for poem writing. \Cref{feedback} in \Cref{sec:appendix} shows some of the feedback provided by experts, including how they found the system helpful in situations such as \textit{writers' block}, and how specific instructions helped them write better. 

\begin{figure}[ht!]
    \centering
    \includegraphics[width=0.42\textwidth]{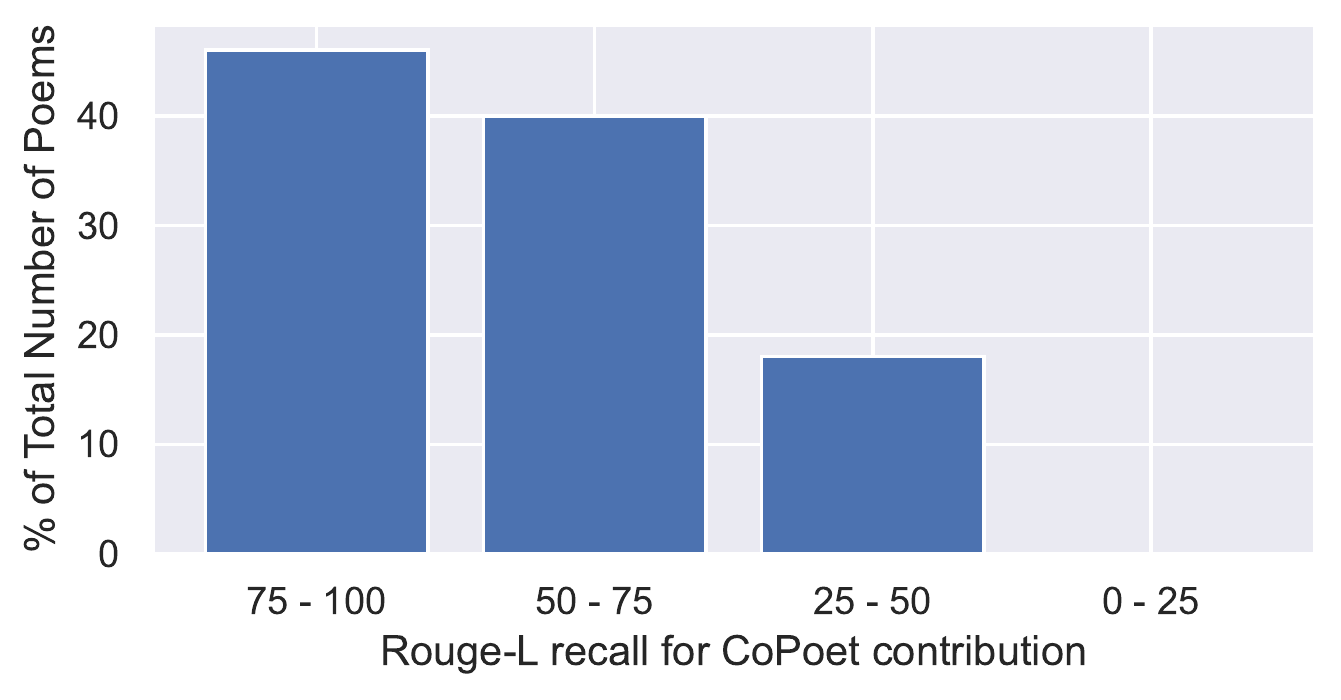}
    \caption{\label{collaboration} 
    Content 
    overlap between sentences of an individual poem and the corresponding model suggestions calculated using Rouge-L recall. Y axis shows the percentage of poems out of 50 while X axis shows the amount of Copoet contribution in terms of Rouge-L.
    }
\end{figure}

\paragraph{What fraction of the poems is written by CoPoet?}
To quantify the contribution of the model, 
we compute the 
proportion of the submitted poems 
that was taken from the model generations. 
We calculate this using the 
Rouge-L recall \cite{lin2004rouge} score of the poem lines with respect to the model suggestions i.e.\ what fraction of the poem is found in the generated output of the model. 
Each verse is greedily matched to a unique model suggestion with the largest overlap. The calculation is described in \Cref{rouge} in Appendix \ref{sec:appendix}.
Figure \ref{collaboration} shows that on average 46\% 
of collaborative poems have a 
Rouge-L recall score greater than 75\%, i.e  75\% of the content in the collaborative poems are obtained from CoPoet suggestions. Additionally, a further 40\% of the 
collaborative poems have more than half of their content (50-75\%) written by the model. This suggests that the majority of the text generated by CoPoet is considered high-quality and usable 
by the expert users. 

\begin{table}[]
\centering
\small
\begin{tabular}{lrr}
\toprule
              & \textbf{Relevant \%} & \textbf{Preferred \%} \\ \midrule
\textbf{Solo}          & 96     & 43        \\ 
\textbf{Collaborative} & 98    & \bf{57}        \\ \bottomrule
\end{tabular}
\caption{\label{turing}Human evaluation of 50 poems written by solo-writers vs 
those written by users with CoPoet.
Workers have a slight 
preference for 
collaborative poems.
}
\end{table}

\paragraph{RQ2: Can CoPoet help users write better poems compared to when they write alone?} \label{betterpoem}
To answer the above question, we compare poems written by the set of experts with and without model help, as detailed in \Cref{fig:copoet_third_party}.  
We are interested in measuring i) whether poems written 
are \textit{relevant}, where an relevant poem is defined as \textit{descriptive interpretation} of the title, i.e.\ it is on-topic. ii) whether poems written by experts with CoPoet are preferred over poems written by solo-writers.

We recruit a total of 49 third-party annotators to compare 
poems written by experts alone to 
those written by experts with CoPoet. They are shown one poem each from a solo-writer and a collaborative writer, both in response to the same title, and requested to label each poem on whether it is \textit{relevant}. Additionally, they 
are asked to choose their preferred poem between the given 
pair 
in terms of coherence, overall quality, and style. Each pair of poems 
is evaluated by 3 distinct annotators. We then aggregate the judgments via \textit{majority voting}. Table \ref{turing} shows that both poems written by solo writers and poems written collaboratively are accurate. 
We are encouraged to see that collaborative poems are  
preferred more than poems written by solo-writers. These findings suggest that CoPoet is a helpful tool for poetry writing 
and instructions act as a useful vehicle for co-creative writing using LLMs.

\begin{table}[ht!]
\centering
\small
\begin{tabular}{lrr}
\toprule
           & \textbf{Preferred \%} & \textbf{Not Preferred \%} \\ \midrule
\textbf{Diversity}  & 63.0      & 37.0          \\ 
\textbf{Rhyme}      & 72.5      & 27.5          \\ 
\textbf{Perplexity} & 55.0      & 45.0          \\ \bottomrule
\end{tabular}
\caption{\label{tab:preference} Analysis of poems preferred by third-party annotators based on (i) rhyme (ii) diversity and (iii) perplexity.
Workers' preference is correlated with the presence of rhyming and vocabulary diversity.
}
\end{table}

\paragraph{Potential Factors for User Preference}
We acknowledge that there is some degree of subjectivity in the user preferences. 
To better understand why a certain poem is preferred by crowd-workers, we investigate whether certain factors correlate with their choices. 
We measure i) \textit{Diversity} (in terms of distinct unigrams) ii) \textit{Presence of Rhyme} (whether there at least one pair of rhyming lines in the poem), and iii) \textit{Perplexity} measured using a pre-trained GPT-2 model for each poem. 
As can be seen in \Cref{tab:preference}, crowd-workers preferred poems that are diverse and have a rhyme scheme 63\% and 72.5\% of the time. From \Cref{fig:inst_count}, we know that our experts tend to use the model to express their ideas by eliciting text from the model that contains specific content but is subject to various constraints (\textit{Topic+Rhyme} and the various \textit{End} instructions). 
Here, we observe that these constraints combined with more diverse vocabulary usage might be contributing to the preference for collaborative poems over solo poems. 

%% file: related.tex
\section{Related Work}

\paragraph{Collaborative Writing}

The key challenge in collaborative writing is to understand user intent so as to provide 
timely and useful suggestions. 
Prior work in story writing 
\cite{Roemmele2015CreativeHA, 10.1145/3172944.3172983} presented 
sentence-level continuations 
at locations specified by a user. 
\citet{akoury-etal-2020-storium, lee2022coauthor} took this a step further providing users with a paragraph of text which they could further edit in story writing and argumentative writing tasks. However, model suggestions of this autocomplete nature were not always helpful, as they often diverged from the user intent \cite{10.1145/3172944.3172983} resulting in only a fraction of generated text being retained \cite{akoury-etal-2020-storium}. Instead of providing a machine-written draft, \citet{padmakumar2021machine} showed that having the model rewrite text only at locations specified by the user results in more helpful suggestions 
in the task of creative image captioning. 

We focus on the task of collaborative poem writing, which adds an additional challenge as useful suggestions need to satisfy several lexical and form constraints (rhyme, meter, sound). Past work for this task has used retrieval to provide suggestions for substitutions at the word and phrase level \cite{Chen2014PoetryOT} or verses that follow different styles \cite{uthus2021augmenting}, but these are unable to dynamically generate novel text. In our work, we utilize large language models to generate text that satisfies the various constraints specified by users, with the added benefit that they can spell out these using natural language instructions. 
Concurrent work has also shown that large language models can help users write scripts and screenplays \cite{mirowski2022co} and longer stories \cite{yang2022re3} by generating text that incorporates structural context via prompt chaining.

\paragraph{Interaction with Users}

Recent work in NLP has highlighted the success of generative large language models as interaction interfaces for the task of creative writing. Finetuning models on tasks verbalised as instructions has shown good generalization to unseen instructions  \cite{wei2021finetuned, sanh2021multitask, mishra2021cross, chung2022scaling}. In our work, we focus on a suite of instructions specific to creative writing and additionally evaluate the instruction-tuning setup with real users who iteratively ask for suggestions in natural language. 

In addition to fine-tuning models on instructions, large language models are also able to generalize to unseen tasks in a few-shot manner when the task is specified as part of the prompt in natural language \cite{ouyang2022training}.  
\citet{reif-etal-2022-recipe} present a prompting method 
which performs 
style transfer 
in a zero-shot or few-shot manner with only a natural language instruction describing the target style without model fine-tuning or exemplars in the target style. 
Unlike most of the recent work that prompts large language models to elicit content \citet{coenen2021wordcraft} frame collaborative writing as a conversation between a human and a LLM-based dialog system and show how the spontaneous utilities of conversation support a variety of interactions. 
More recently \citet{mishra2022help} propose a prompting strategy where they ask GPT3 specific questions about mood, tone, occasion, or theme for the task of poem generation by using GPT3 as an interaction interface. 


%% file: conclusion.tex
\section{Conclusion}
In this work, we present \textit{CoPoet}, a collaborative poetry writing system that is controlled by user instructions that specify 
the attributes of any desired text. Our system is built upon a language model fine-tuned on a diverse collection of instructions for poetry writing. Empirical results show that our model is not only competitive with publicly available LLMs trained on instructions (InstructGPT), but also capable of satisfying unseen compositional instructions. A further study with 15 qualified crowd-workers shows that users successfully write poems with CoPoet on diverse topics, which are also preferred by third-party evaluators over poems written by solo-writers. 
These results show 
that language models acting as writing assistants are capable of understanding user intents and collaborating with them to improve the final outcome, potentially makes a challenging task such as poem writing more accessible to users.

Going forward we hope to extend our research to 
more challenging instructions such as converting longer content planning tasks into the instruction tuning setup to assist users with longer story writing. To provide more robust assistance, we also hope to study how to train models that generalize better to completely unseen instructions. Finally, we intend to more holistically study the problem of co-creative writing by not just examining how to train better assistive models but also how to design effective user interfaces for end users. 

\section*{Limitations}
\paragraph{Noisy Training Data}
We note that our dataset is self-supervised and we use various tools to align lines of poetry from various sources (\Cref{instrsource}) to templated instructions. There might be small errors in the training data such as spelling mistakes in the lines of poetry (an example from our dataset to showcase this is the line ``Lay silently \textit{burid} side by side'') or slightly convoluted instructions (an example instruction to highlight this is ``Write a poetic sentence that speaks of \textit{nights grow shorter}''). However, each example in the various test sets (\Cref{sec:test_sets}) was manually verified by the authors of this work. 

\paragraph{Test Set Size}
Another potential concern is the size of the test sets which were small as each instruction in these was verified by the authors. We provide confidence intervals on the model success rates to mitigate this in \Cref{sec:automatic_evaluation}. 

\paragraph{Design of the User Interface}
Our user interface presents templates of instructions to users at the point when they query the model for assistance (\Cref{fig:humanai}). This primes the users to write instructions similar to the templates---
almost all the instructions used by the crowdworkers belonged to the templates provided in the interface (or novel combinations of these). In this work, we did not perform an extensive comparison of different interface designs which could influence the interaction. We further discuss some of the design choices about the user interface in \Cref{sec:user_interface}.




\section*{Ethics Statement}
Although we use language models trained on data collected from the Web, which have been shown to have issues with gender bias and abusive language, the inductive bias of our models should limit inadvertent negative impacts. Unlike model variants such as GPT, T5 is a conditional language model, which provides more control of the generated output. Our poetic parallel corpora are unlikely to contain toxic text and are manually inspected by the authors.Technological advances in text generation have had both positive and negative effects. However, interactive, human-in-the-loop generative systems designed especially for literary or poetic text generation such as ours might speed up literary professional's work and make it more enjoyable. We believe that machine generation of poetic text will not lead to the exclusion of human poets. Rather, it will increase human-machine interaction and continue to enhance human performance.

In order to ensure that there are no privacy issues for our train and validation splits, the poems were broken down line by line and shuffled randomly.They do not contain any metadata and as such cannot reproduce the creative value of the original poems.

\paragraph{Appropriate Remuneration of Crowd-workers}
For all our tasks, we recruit from a pool of crowd-workers in the USA with a minimum of 95\% HIT success rate. 
To complete the human evaluation of model outputs satisfying instructions (\Cref{sec:human_evaluation}), a crowdworker has to read an instruction and two lines in response to it and answer a total of 5 questions. On average, this takes slightly less than two minutes, so we set the payment to \$0.50 per HIT. For the writing tasks (solo and collaborative, \Cref{co-poet-description}), on average our users take 10 minutes to write a poem, so we set the payment of \$2.50 for each HIT. We also reward writers \$0.50 per poem on submission of poems deemed \textit{relevant} or a relevant interpretation of the title, per the definition in \Cref{co-poet-description}. Over 95\% of the poems submitted received a bonus (\Cref{turing}). Finally, for the judging task of comparing solo-writers and collaborative writers, crowdworkers have to read two poems and answer 3 questions, which takes on average 1 minute, so we set the payment to \$0.25 per HIT. All of these amounts were calculated according an hourly rate of $15\$$ per hour.     

%% file: appendix.tex
\section{Appendix}
\label{sec:appendix}

\subsection{Creation of Instructions}

To create instructions for a particular ``Subject" we detect all possible noun phrases from an individual poetic sentence and create a natural language instruction for each of them using the template describe in Table \ref{trainins}. For ``End" we fill the respective instruction template with the ending word in a sentence. For the``Rhyme" instruction we first find all rhyming words for the ending word in a sentence using the CMU Pronouncing Dictionary \footnote{https://pypi.org/project/pronouncing/} and then fill the instruction template with a random rhyming word to ensure diversity. For the ``Next Sentence" we fill the instruction template with its previous context sentence from any given poetry. To create ``Metaphor" instruction we crawl websites for outputs of the form ``NP1 is NP2" and fill NP1 in the template. A `Simile" usually consists of two noun phrases typically a Subject and an Object with an usual syntax ``NP1 is like NP2". We fill the Subject NP1 in the instruction template and manually edit it by expert humans for any inconsistencies. It should be noted that both output quality and instructions for Simile and Metaphors are manually inspected and agreed upon by two expert humans and only examples with full agreement are kept in the data. To create the instruction for `Haiku' we need to fill the template with its title which is not always readily available. Hence we use YAKE \cite{campos2018yake,CAMPOS2020257}, an unsupervised automatic keyword extraction method for selecting salient words from the Haiku that serves as its title. For Onomatopoeia we compile a lexicon containing words \footnote{https://kathytemean.wordpress.com/2009/12/29/onomatopoeia-word-list/} representing them and then filter out sentences with any noun subject containing a word from the lexicon. 
\begin{table}[]
\centering
\small
\renewcommand{\arraystretch}{1.25}
\begin{tabular}{|l|l|l|}
\hline
Instruction Type                                                              & Source & Stats                                                                           \\ \hline\hline
\begin{tabular}[c]{@{}l@{}} Lexical Constraint\end{tabular} &  \begin{tabular}[c]{@{}l@{}}Poetry Translation Corpus\\ \citet{chakrabarty-etal-2021-dont}  \end{tabular} & 94.5\%                                                                                 \\ \hline\hline
\begin{tabular}[c]{@{}l@{}} Continuation\end{tabular} &  \begin{tabular}[c]{@{}l@{}}Poetry Translation Corpus\\ \citet{chakrabarty-etal-2021-dont}  \end{tabular} & 3.18\%                                                                                 \\ \hline\hline
Rhetorical Devices                                                            & \begin{tabular}[c]{@{}l@{}}r/OCPoetry , r/Poetry\\ Gutenberg \citet{jacobs2018gutenberg},\\ DMDMQ \footnote{https://www.drmardy.com/dmdmq/} \end{tabular}  & 1.12\%\\ \hline\hline
Haiku                                                                         & r/Haiku & 1.14\%                                                                          \\ \hline
\end{tabular}
\caption{\label{instrsource}Instruction Types along with the source from where the data is collected.}
\end{table}

\begin{table}[]
\small
\centering
\renewcommand{\arraystretch}{1.15}
\begin{tabular}{|l|}
\hline
\begin{tabular}[c]{@{}l@{}}The AI is very competent and helpful,it's enjoyable\\ to work with it.\end{tabular}                                                 \\ \hline\hline
\begin{tabular}[c]{@{}l@{}}I think it works very fine and I wish I had this whenever\\ I had writer's block.\end{tabular}                                     \\ \hline\hline
\begin{tabular}[c]{@{}l@{}}The best part of the tool is getting help with words at the \\ end of a sentence and then being able to build off that.\end{tabular} \\ \hline
\end{tabular}
\caption{\label{feedback}Some of the feedback from experts on the helpfulness of using our CoPoet system.}
\end{table}

\begin{algorithm}[h!]
\small
\caption{Algorithm to compute how much of final submitted poem is written using model output.}
\label{rouge}
\begin{algorithmic}
\State Let $S = \{i,o\}$ be the set of all instructions requested by the experts and corresponding model outputs for a \textbf{single poem};
\State $sum\_RL = 0 ; n = num\_lines(poem)$\\
\ForEach{line $l \in  poem$}
\State $max\_rouge = -1$
\State $max\_tuple = None$
\ForEach{$(i',o') \in S$}
\State $rouge\_score = RougeL(o',l)$
\If{$rouge\_score > max\_rouge:$}
\State $max\_rouge = rouge\_score$
\State $max\_tuple = (i',o')$
\EndIf
\EndFor
\State $sum\_RL = sum\_RL + max\_rouge$
\State $S' = S$ ; $S' = S' - max\_tuple$; $S = S'$
\EndFor
\State $poem\_RL = sum\_RL/n$
\end{algorithmic}
\end{algorithm}

\section{Poems from User Study}
\label{sec:poem_examples}
\begin{figure}
    \centering
    \includegraphics[width=0.475\textwidth]{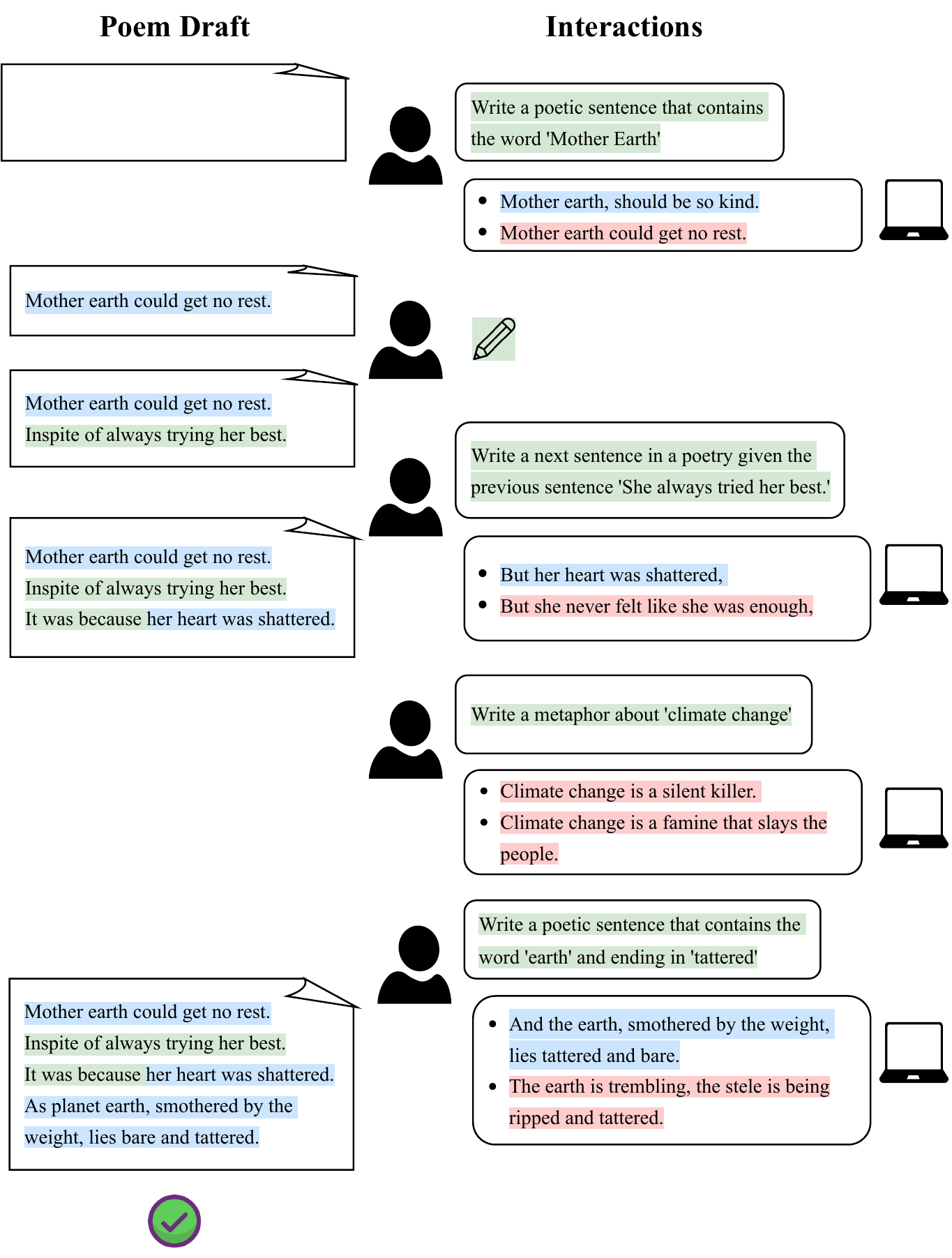}
    \caption{Poem entitled `Climate Change' written in collaboration with CoPoet.}
    \label{fig:copoet_climate_change}
\end{figure}

\begin{figure}
    \centering
    \includegraphics[width=0.475\textwidth]{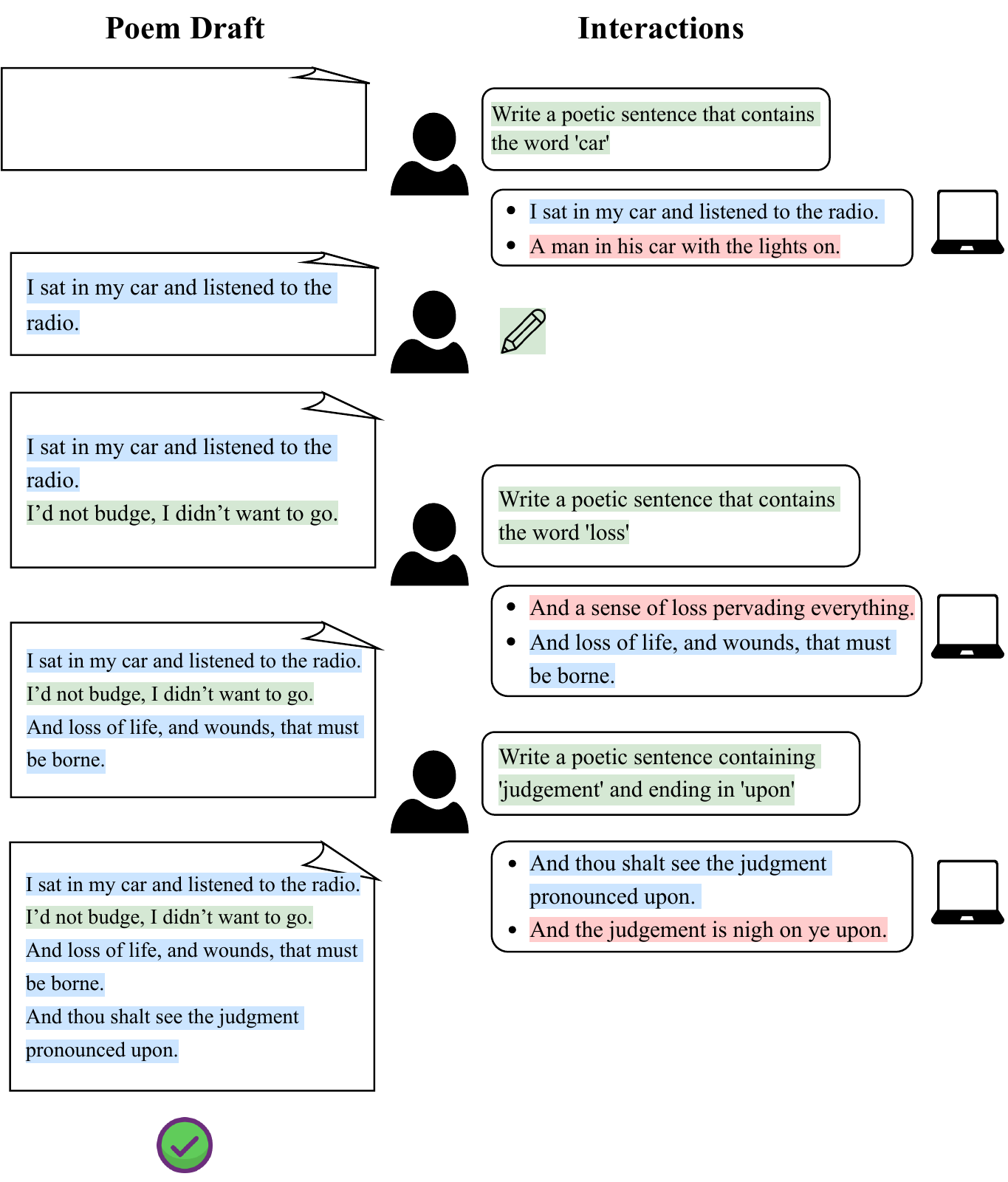}
    \caption{Poem entitled `Courthouse Parking Lot' written in collaboration with CoPoet.}
    \label{fig:copoet_courthouse_parking_lot}
\end{figure}

\begin{figure}
    \centering
    \includegraphics[width=0.475\textwidth]{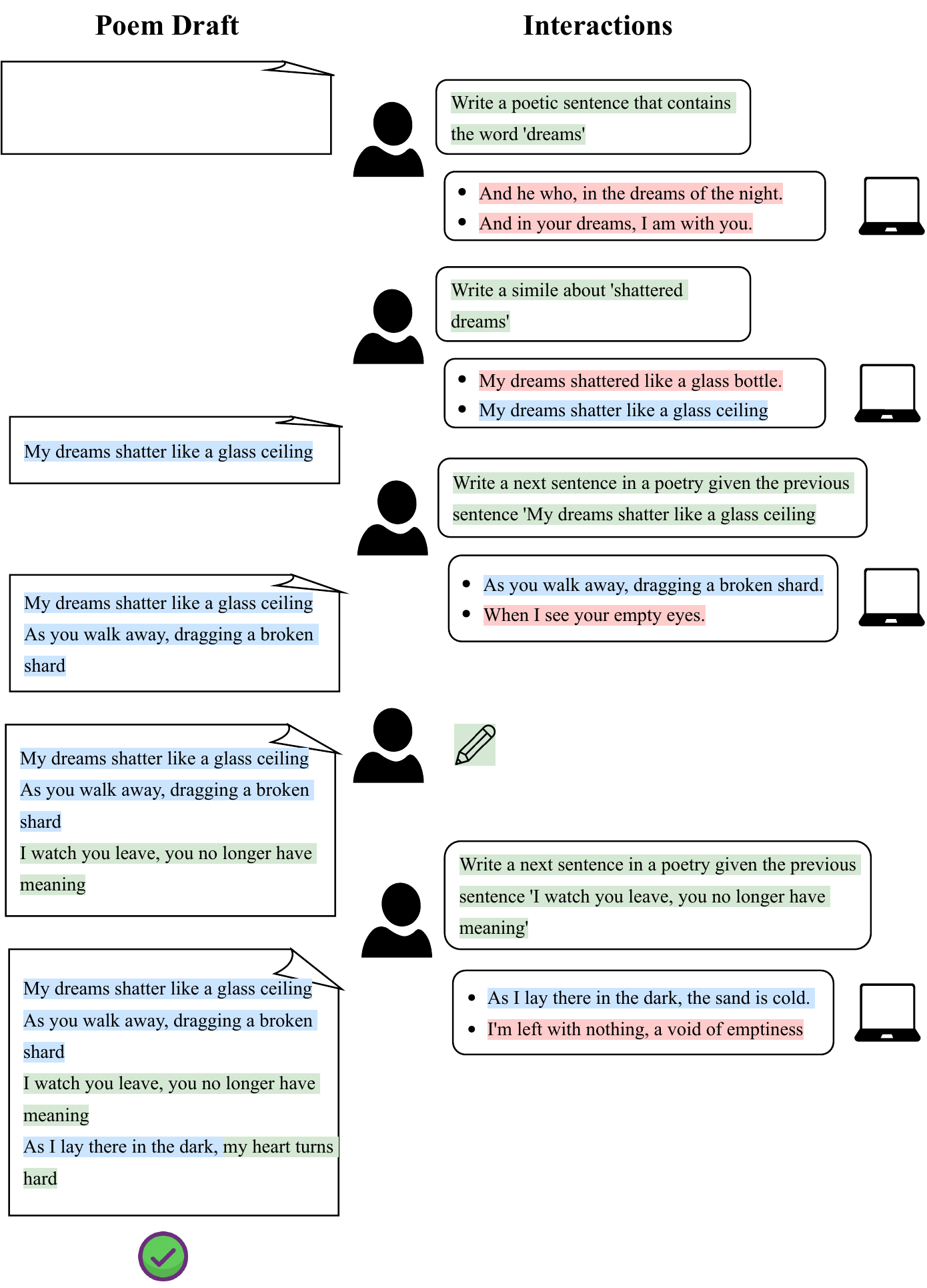}
    \caption{Poem entitled `Glass Ceilings' written in collaboration with CoPoet.}
    \label{fig:copoet_glass_ceilings}
\end{figure}

\begin{figure}
    \centering
    \includegraphics[width=0.475\textwidth]{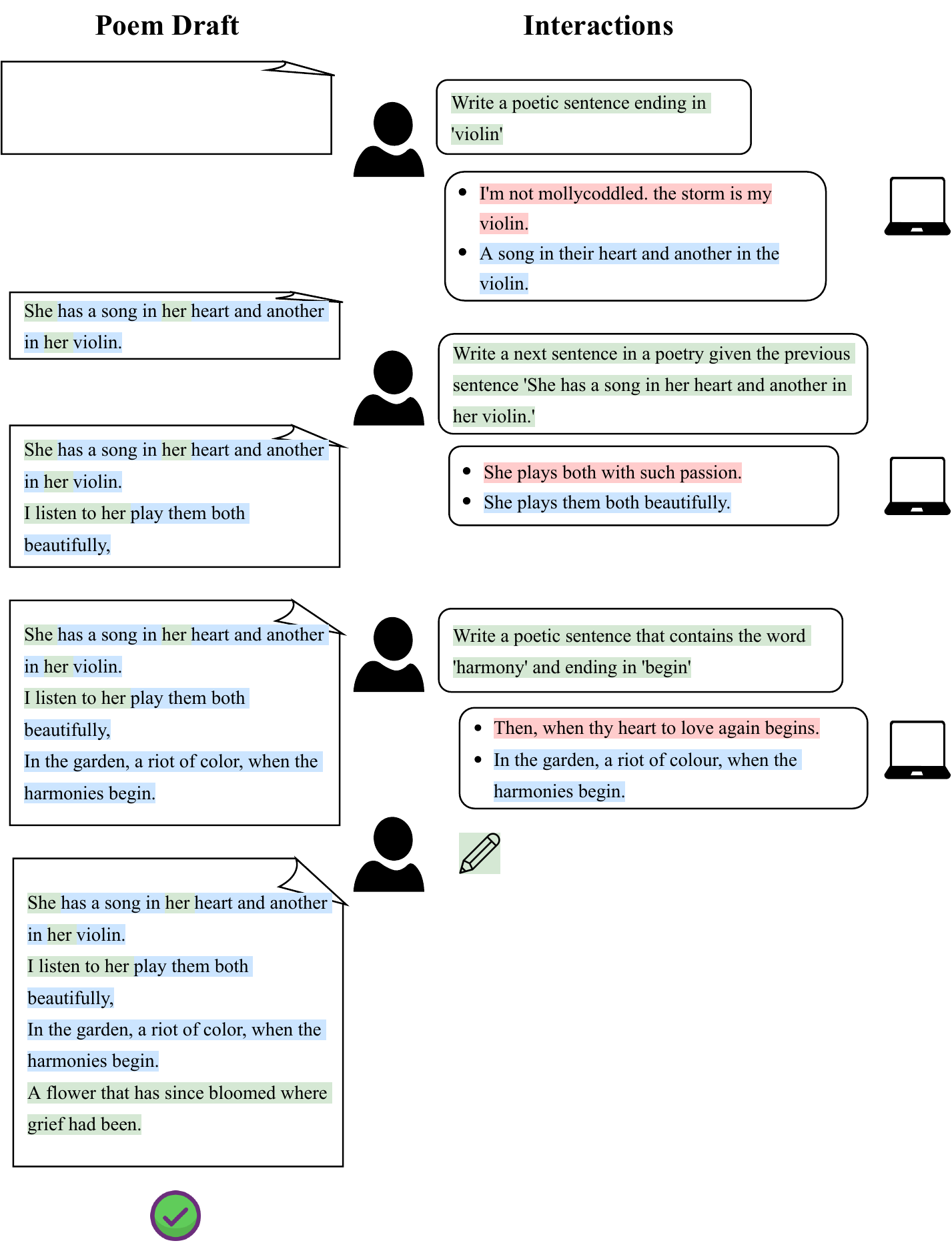}
    \caption{Poem entitled `Petal Melody' written in collaboration with CoPoet.}
    \label{fig:copoet_petal_melody}
\end{figure}

\begin{figure}[t]
    \centering
    \includegraphics[scale=0.38]{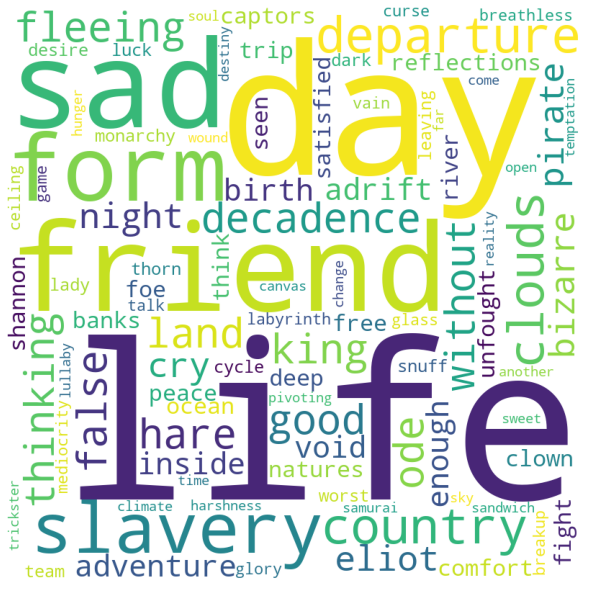}
    \caption{\label{fig:wcloud} Word cloud of different concepts from titles of Poetry.}
\end{figure}

We attach further examples of poetry written in collaboration with CoPoet in \Cref{fig:copoet_climate_change,fig:copoet_courthouse_parking_lot,fig:copoet_glass_ceilings,fig:copoet_petal_melody}. 
These include instances where the user selects none of the options presented to them (\Cref{fig:copoet_climate_change,fig:copoet_glass_ceilings}) and highly intertwined collaboration where the user frequently rewrites model output (\Cref{fig:copoet_petal_melody}). Additionally, \Cref{fig:wcloud} is a word cloud of the titles of all the poems written by the users.

\section{User Interface}
\label{sec:user_interface}
\begin{figure*}[t]
    \centering
    \fbox{\includegraphics[width=\textwidth]{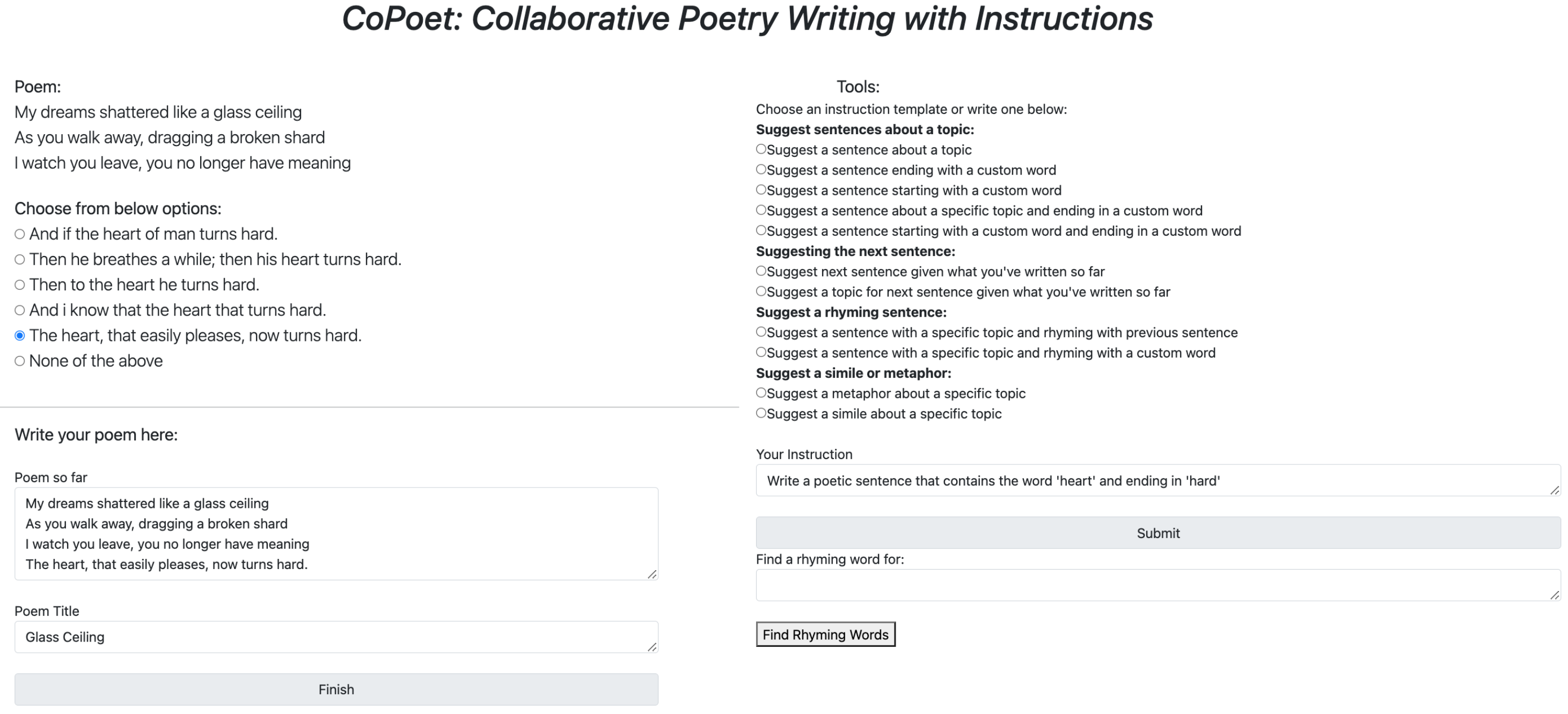}}
    \caption{\label{fig:humanai} Snapshot of CoPoet: Collaborative Poetry Writing with Instructions}
    \label{fig:architecture_copoet}
\end{figure*}

A snapshot of our interface during the user study can be found in \Cref{fig:architecture_copoet}. The user is presented with a text box to edit their poem draft along with a dialog box to query the model.
From an initial pilot, we observed that some users were not able to effectively write instructions. As a result, we chose to provide instruction templates as part of the interface in the form of radio buttons (\Cref{fig:humanai}). This was aimed at informing novice writers of the kind of instructions that elicit creative lines of text (rhymes, metaphors, etc.) from the model which they can then use to write better poems (which we noted in some feedback obtained from crowdworkers).
However, this also primes the users to write instructions similar to the templates---
almost all the instructions used by the crowdworkers belonged to the templates provided in the interface (or novel combinations of these).   
In this work, we did not perform an extensive comparison of different interface designs which could influence the effectiveness of human-AI collaboration. Our main goal is to design and test the instruction tuning setup specifically for the poetry writing task, which was why we chose to retain the interface design with the templates. 
The user interface ensures that most of the queries to the model follow the same templates which need not be the case in deployment. From very preliminary experiments, we see that InstructGPT3 outperforms our fine-tuned model on completely unseen instructions, and we intend to investigate this setting in detail going forward. We believe that as we provide users with greater flexibility in how to use the interface, the associated model must be able to respond robustly to the edge cases that users might provide and hence needs extensive rounds of piloting prior to deployment.


